\documentclass[preprint,12pt]{elsarticle}

\usepackage{amssymb}
\usepackage{amsmath}

\usepackage{soul}
\usepackage{xcolor}

\usepackage{pgfplots}
\pgfplotsset{compat=1.18} 
\usepgfplotslibrary{statistics} 
\usepackage{pgfplotstable} 
\usetikzlibrary{patterns, positioning}

\usepackage{amsfonts}
\usepackage{booktabs}
\usepackage{multirow}
\usepackage{graphicx}
\usepackage{pdflscape}

\usepackage{amsmath,amssymb,amsfonts}

\usepackage{textcomp}
\usepackage{algorithm}
\usepackage{algpseudocode}

\usepackage{fancyhdr}

\usepackage[hyphens]{url}

\usepackage{pdflscape} 

\journal{Telematics and Informatics Reports}

\begin{document}

\begin{frontmatter}



\title{Privacy-Preserving in Connected and Autonomous Vehicles Through Vision to Text Transformation}


\author[label1]{Abdolazim Rezaei}
\author[label1]{Mehdi Sookhak} 
\author[label4]{Ahmad Patooghy}
\author[label2]{Shahab S. Band} 
\author[label3]{Amir Mosavi} 

\affiliation[label1]{organization={Department of Computre Science, Texas A\&M University Corpus Christi},
            addressline={6300 Ocean Dr}, 
            city={Corpus Christi},
            postcode={78412}, 
            state={Texas},
            country={USA}
            }
\affiliation[label2]{organization={Department of Information Management, International Graduate School of Artificial Intelligence, National Yunlin University of Science and Technology},
            city={Douliu},
            postcode={78412}, 
            country={Taiwan}
            }            
\affiliation[label3]{organization={Institute of the Information Society, Ludovika University of Public Service},
            city={Budapest},
            postcode={78412}, 
            country={Hungary}
            }
\affiliation[label4]{organization={North Carolina A\&T State University},
            addressline={601 E Market}, 
            city={Greensboro},
            postcode={27411}, 
            country={USA}
            }            
\begin{abstract}
Intelligent Transportation Systems (ITS) rely on a variety of devices that frequently process privacy-sensitive data. 
Roadside units are important because they use AI-equipped cameras to detect traffic violations in Connected and Autonomous Vehicles (CAV).
However, although the interior of a vehicle is generally considered a private space, the privacy risks associated with captured imagery remain a major concern, as such data can be misused for identity theft, profiling, or unauthorized commercial purposes. 
Methods like face blurring reduce privacy risks, however individuals' privacy can still be compromised.
This paper introduces a novel privacy-preserving framework that leverages feedback-based reinforcement learning (RL) and vision-language models (VLMs) to protect sensitive visual information captured by AIE cameras. 
The proposed idea transforms images into textual descriptions using an innovative method while the main scene details are preserved and protects privacy.
A hierarchical RL strategy is employed to iteratively refine the generated text, enhancing both semantic accuracy and privacy. 
Unlike prior captioning-based methods, our model incorporates an iterative reinforcement-learning cycle with external knowledge feedback which progressively refines privacy-aware text. In addition to qualitative textual metric evaluations, the privacy-based metrics demonstrate significant improvements in privacy preservation where SSIM, PSNR, MSE, and SRRA values obtained using the proposed method on two different datasets outperform other methods.
\end{abstract}




\end{frontmatter}

\section{Introduction}
Privacy in Intelligent Transportation Systems (ITS) has become a major concern in recent years, especially as a growing number of electronic devices are integrated into these vehicles. These systems transfer vehicle data to machine learning (ML)-based image processing pipelines for tasks such as image classification, object detection, face detection and recognition, image enhancement and restoration, and augmented reality. Among these technologies, Artificial Intelligence Equipped (AIE) cameras, which are increasingly deployed at junctions and traffic lights—are now integrated with CAV systems to monitor driver behavior (e.g., mobile phone use and seatbelt compliance) and to capture images for traffic monitoring, violation detection, and safety assessment, all aimed at enhancing public safety.

\begin{figure}
\centerline{\includegraphics[width=0.65\linewidth]{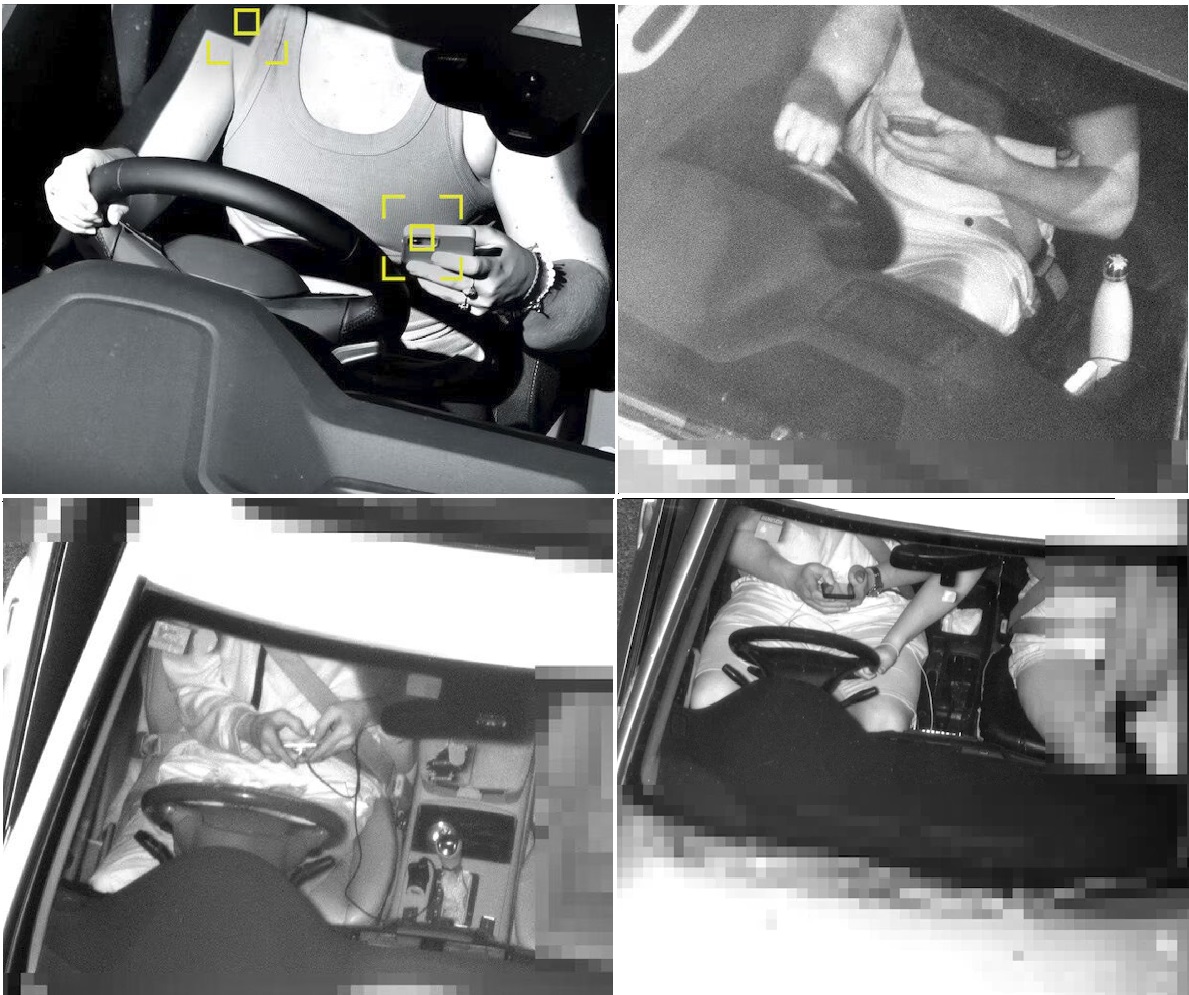}}
\caption{Captured images are used to detect violations; however, as it is obvious, the privacy is compromised.}
\label{fig:privacy}
\end{figure}

Despite the financial and operational success of AIE camera systems, their deployment has raised serious privacy concerns. For example, in Queensland, Australia, an AIE camera initiative has generated approximately \$419.8 million in revenue since 2021 \cite{queensland1, queensland2}, however potential privacy related issues arise from captured images. 
As illustrated in Figure~\ref{fig:privacy}, the interior of vehicles—widely regarded as a private space—can be inadvertently exposed, raising legal and ethical concerns. Similar issues have been reported in U.S. states such as California and Maryland, where AIE cameras have been deployed despite ongoing legal challenges \cite{theSun2024trafficAI}. 
These deployments highlight growing concerns about visual data privacy in connected and autonomous vehicles. The issue includes not only images but also GPS information, biometric data, and audio recordings \cite{liu2021privacy}.

According to recent studies \cite{boovaraghavan2024kirigami, liu2021privacy}, compromising the privacy of such data can result in consequences such as identity theft, unauthorized surveillance, user profiling, and commercial exploitation. Xie et al. \cite{xie2022privacy} emphasize that the vehicle interior is considered as personal space and must be protected. While techniques such as blurring and obfuscation are commonly used to mask sensitive content, they are usually inaccurate and may fail to fully conceal private regions. Additionally, recent advances in AI-based reconstruction attacks show serious threats, enabling adversaries to reverse these privacy measures and reveal the original content.

Considering the limitations of existing privacy-preserving techniques, such as inaccurate blurring and vulnerability to AI-based reconstruction attacks, there is a need for more robust solutions. In response, this study proposes a novel framework that transforms visual data into descriptive text using advanced VLMs, thereby avoiding the direct exposure of image content.

Unlike traditional image captioning or single-pass text generation, our approach employs a hierarchical RL loop that iteratively refines semantic richness through external feedback from Retrieval Augmented Generation (RAG). Each iteration improves the contextual precision of descriptions beyond what static captioning or one-time VLM generation can achieve.

The core idea of the proposed model is to leverage VLM techniques to generate descriptive text from collected images which captures all necessary information without revealing the actual visual content. To enhance the quality and relevance of the generated descriptions, we introduce a feedback-based Reinforcement Learning (RL) framework that operates in parallel with a RAG mechanism. RAG plays a key role in optimizing prompt selection by double-checking and refining the prompts used during RL training. To the best of our knowledge, this is the first attempt to combine RL and VLM for transforming visual data into text as a means of privacy preservation.

\noindent The core contributions of this study are as follows:
\vspace{-4pt}
\begin{enumerate}
    \item Development of an iterative image-to-text transformation architecture, which generates semantically rich and privacy-aware descriptive text from visual data. 
    \item Design of a hierarchical RL mechanism that refines both  prompts and generated outputs in multiple stages.
    \item Integration of a feedback-based learning loop using RAG to validate and optimize prompt selection, thereby improving the relevance and accuracy of the generated text.
\end{enumerate}

The rest of this paper is organized as follows. Related Work reviews the existing literature and background methods. Proposed Model introduces the RL-VLM framework. Results and Discussions describes the experimental setup and presents the results. Finally, Conclusion and Future Work summarizes the key findings and outlines directions for future research.

A list of acronyms used in this article is provided in Table \ref{table:acronyms}.

\begin{table}
\caption{List of acronyms used in this article.}
\centering
\begin{tabular}{|c|c|c|}
\hline
\textbf{Acronym} & \textbf{Expression} \\
\hline \hline
AIE     & AI-Equipped \\ \hline

CAV(s)  & Connected and Autonomous Vehicle(s) \\ \hline
DP      & Differential Privacy \\ \hline
FL      & Federated Learning \\ \hline
IoT     & Internet of Things \\ \hline
ITS     & Intelligent Transportation Systems \\ \hline
LLM     & Large Language Model \\ \hline
MIPS    & Maximum Inner Product Search \\ \hline
MSE     & Mean Squared Error \\ \hline
PPO     & Proximal Policy Optimization \\ \hline
RAG     & Retrieval Augmented Generation \\ \hline
SBERT   & Sentence-BERT \\ \hline
RL      & Reinforcement Learning \\ \hline
SSIM    & Structural Similarity Index Measure \\ \hline
VLM(s)  & Vision Language Model(s) \\ \hline
\end{tabular}
\label{table:acronyms}
\end{table}

\section{Background}
Feedback-based RL is a specialized form of reinforcement learning where an agent learns through continuous feedback from the environment, but with an added emphasis on using richer, more informative feedback signals. Traditional RL typically relies on a simple reward signal, which might be sparse or delayed, meaning that the agent only receives feedback in the form of a reward or punishment based on the cumulative outcome of its actions. However, feedback-based RL introduces the concept of providing more nuanced or structured feedback, such as expert annotations, preferences, or evaluations from human users, to guide the agent toward better decision-making. This feedback could come from various sources and may not be limited to just numerical rewards; instead, it might include qualitative assessments, rankings, corrections, or other types of informative signals that help the agent understand what constitutes a "good" or "bad" action.

\subsection{Mechanisms in Feedback-Based RL}
In feedback-based RL, the agent still interacts with the environment by observing states, taking actions, and receiving feedback. However, the feedback may be more sophisticated and it often helps the agent to understand not just the outcome of its actions but also how to improve its decisions. For example, in a traditional RL setting, if an agent is learning to play a video game then it might only receive feedback when it wins or loses the game which can be sparse during the learning process. In contrast, feedback-based RL would involve providing the agent with more feedback such as pointers on specific decisions it makes, hints from an expert player, or even visual and verbal cues from the environment about how well it is progressing.

One of the key benefits of feedback-based RL is that it accelerates the learning process by reducing the reliance on sparse or delayed rewards. In environments where actions only produce feedback after a long sequence of decisions, traditional RL can struggle because the agent finds it difficult to attribute success or failure to specific actions. By incorporating more frequent and informative feedback, the agent is able to learn more efficiently. The feedback mechanism would be through the following ways:

\begin{enumerate}
    \item \textbf{Human Feedback}: In some systems, humans provide feedback to guide the agent’s learning. This feedback can be explicit such as ranking actions or implicit feedbacks such as preference-based feedback where humans select actions they prefer.

    \item \textbf{Expert Annotations}: In certain domains like healthcare or robotics, expert annotations can be used to guide agents. Experts can provide corrective actions or evaluations which help the agent to better understand the situation in which it is dealing with. 

    \item \textbf{Preference Learning}: In some cases, feedback comes in the form of preferences where an agent is given feedback comparing two different sequences of actions or outcomes which helps it understand which sequence leads to better outcomes.

    \item \textbf{Corrective Feedback}: In robotics or control systems, an expert might directly intervene to correct the agent’s actions. This immediate, direct feedback helps the agent to refine the learning more quickly than waiting for the final outcome of a task.

\end{enumerate}
Feedback-based RL is particularly useful in domains where high-quality or real-time feedback is available but learning from sparse, long-term rewards would be inefficient. By leveraging external feedback sources, the agent can often reach optimal policies faster and with less exploration.

\subsection{Types of Feedback in Feedback-Based RL}
Feedback-based RL can utilize different types of feedback beyond just rewards and penalties. One commonly used feedback type is \textbf{human-in-the-loop feedback} in which a human provides guidance to the agent in real time. This type of feedback can be useful in domains where an expert can offer insights that are not easily encoded into traditional reward functions. For example, in healthcare, a doctor can provide feedback on the quality of a treatment process devised by an RL agent, adjusting the agent’s decisions based on nuanced clinical knowledge. This expert feedback is better than a simple reward and provides the agent with actionable information about how to improve its policy.

Another form of feedback in this context is \textbf{preference-based feedback}, where an agent learns from feedback which involves ranking or comparing actions or outcomes. Instead of receiving a scalar reward for each action, the agent is presented with pairs of outcomes and asked which one is better. This is a more nuanced form of learning because the agent can start to understand relative preferences rather than learning purely from binary success or failure signals. This method has been particularly successful in applications like recommendation systems in which user's preferences are expressed in comparative rather than absolute terms.

In addition, feedback can also be \textbf{interactive}, where an agent learns from an interaction with users or experts. This interaction allows the agent to clarify its understanding of the task or the environment and to receive specific feedback that helps it learn faster. For example, in robotics, an agent might ask a human operator for clarification when it encounters an unfamiliar object or a difficult task. By integrating feedback in real-time, the agent can improve its learning process dynamically and adjust its actions accordingly.

\subsection{Applications of Feedback-Based RL}
Feedback-based RL has numerous applications across different fields, particularly in areas where obtaining detailed feedback is possible and beneficial. One prominent application is in robotics, where agents can learn complex behaviors by receiving corrective feedback from human operators or experts. For example, a robot might be trained to grasp objects more effectively through a combination of expert guidance and real-time feedback, helping the agent refine its motor control skills.

In gaming, feedback-based RL has been used to improve agents that learn from human play or expert demonstrations. 
For instance, in games like StarCraft II or Dota 2, RL agents have been shown to perform better when they are guided by human experts who provide feedback on strategy, timing, and decision-making. 
By learning from expert strategies and receiving feedback on intermediate actions, the agents can achieve higher performance more quickly than through traditional RL methods alone.

Another area in which feedback-based RL excels is in recommendation systems, such as those used by streaming services or e-commerce platforms. Here, feedback comes in the form of user preferences or ratings, and the agent learns to optimize its recommendations based on what users prefer. For instance, instead of just presenting a set of items and learning from whether the user clicks or not, feedback-based RL can use feedback which users rank options or express preferences between different items. This feedback helps the agent understand more about the user’s underlying preferences, leading to better and more personalized recommendations.

In healthcare, feedback-based RL is used in personalized treatment planning. Doctors can provide feedback on the effectiveness of certain treatment actions, guiding the agent in optimizing treatment plans that are tailored to individual patients. By incorporating expert clinical knowledge through feedback, RL agents can learn to make more informed decisions, particularly in complex and high-stakes environments.

\subsection{RL-based Captioning vs. Feedback-Driven Iterative Description}
Traditional RL-based captioning methods focus primarily on optimizing a language model to generate a single caption that maximizes a linguistic or perceptual reward such as BLEU, CIDEr, or semantic similarity—based on one round of interaction between the RL agent and the captioning model. In these approaches, reinforcement learning serves mainly as a post-processing optimization step, fine-tuning captions to align with ground-truth descriptions. However, these models lack awareness of privacy, contextual feedback, or iterative refinement; once the caption is produced, the process ends without feedback-based adjustment or external evaluation of privacy preservation.

In contrast, the proposed feedback-driven iterative description framework extends beyond static RL captioning by introducing a multi-stage refinement loop where each cycle learns from both internal rewards (semantic similarity via SBERT) and external knowledge-based feedback (via RAG). This dual feedback mechanism enables the model to improve description quality meanwhile privacy is preserved across iterations. 
Instead of producing a single caption, the model evolves its descriptions over multiple rounds—filtering redundant content, enhancing semantic density, and ensures privacy preservation dynamically. This approach transforms RL from a one-time caption optimizer into a continuously adaptive reasoning agent.

\section{Related Work}

\label{Sec:related} 
Privacy in the context of CAVs is a complex term with multiple implications. For example, the authors in \cite{ghaziamin2024privacy} focus on counting passengers at bus station without showing their faces. In \cite{he2024multi}, the goal is to collect traffic data while protecting individuals' privacy. Another study by Zhang et al. \cite{zhang2023privacy} looks inside the vehicle to count passengers, also considering privacy concerns. Therefore, it can be said that the definition of privacy can vary based on the situation. 

There have been many privacy preservation studies where researchers have proposed different ways to manipulate parts of the target image. 
For example, some authors propose a face masking method to issue tickets. In a different way, authors in \cite{popescu2022obfuscation} focus on image obfuscation to protect privacy. Face obfuscation, masking, fogging, and replacements are the most common methods to avoid compromising privacy of images \cite{laishram2025toward, huang2024identity}.
However, existing methods are not capable to ensure privacy preservation. For example, Jian et al. in \cite{jiang2023dartblur} explain how these methods are likely vulnerable to some AI methods such as adversarial attacks where attackers can recover identities. In a similar study \cite{ye2024securereid}, the author discusses the vulnerability of visual information hiding methods that are susceptible to adversarial attacks. 
In the Queensland case \cite{queensland1, queensland2}, the goal was to detect driving violations, but parts of the driver's body were still visible even though partial fogging is applied, and this raises additional privacy concerns.
In a similar case, The State of Maryland uses AIE roadside cameras to detect traffic violations \cite{theSun2024trafficAI}. These cameras try to blur the driver’s face, but this does not fully guarantee visual privacy. In fact, a major problem with these methods is that they may incorrectly target parts of the image that involve privacy concerns \cite{xiao2024privacy, xie2022privacy}.

Privacy in the context of CAVs is a complex term with multiple implications. For example, the authors in \cite{ghaziamin2024privacy} focus on counting passengers at bus station without showing their faces. 
In \cite{he2024multi}, the goal is to collect traffic data while protecting individuals' privacy. Another study by Zhang et al. \cite{zhang2023privacy} looks inside the vehicle to count passengers, also considering privacy concerns. Therefore, it can be said that the definition of privacy can vary based on the situation. 
There have been many privacy preservation studies where researchers have proposed different ways to manipulate parts of the target image. 
For example, some authors propose a face masking method to issue tickets. In a different way, authors in \cite{popescu2022obfuscation} focus on image obfuscation to protect privacy. Face obfuscation, masking, fogging, and replacements are the most common methods to avoid compromising privacy of images \cite{laishram2025toward, huang2024identity}.

Popescu et al. \cite{theSun2024trafficAI} proposed an obfuscation algorithm that combines a Variational Autoencoder (VAE) with random non-bijective pixel intensity mapping to protect sensitive content in medical images while maintaining sufficient fidelity for deep learning–based analysis. Their approach balances the privacy–accuracy trade-off by preventing human and AI-based reconstruction while enabling model training on obfuscated data. This method demonstrated strong performance for privacy-preserving classification in medical imaging, highlighting the potential of hybrid generative–statistical obfuscation strategies for sensitive visual data.

Jiang et al. \cite{jiang2023dartblur} introduced DartBlur, a de-artifact blurring framework that uses a U-Net architecture and adversarial optimization to anonymize facial data while suppressing detection artifacts. 
Unlike traditional Gaussian blurring, DartBlur preserves operational fidelity and post-hoc fidelity which ensures consistent model performance on blurred and clean data. The authors also introduced multiple training objectives to enhance review convenience, accessibility and artifact suppression in order to represent a significant advancement in learnable blur-based privacy preservation for vision models.

Huang et al. \cite{huang2024identity} presented a dual surrogate generative model framework for identity-preserving face swapping that introduces explicit supervision via proxy-paired data. Their Credible Supervision Completion via Surrogates system creates synthetic source–target pairs using two surrogate networks to maintain identity features while modifying non-identity attributes. This dual-surrogate strategy improves visual fidelity and identity consistency compared to prior GAN or latent-based models which establishes a new direction for privacy-aware face synthesis. 

Ghaziamin et al. \cite{ghaziamin2024privacy} developed a privacy-preserving passenger counting system using overhead fisheye cameras and NVIDIA Jetson edge devices which enables real-time inference without cloud transmission. By performing all computations locally, the approach ensures user anonymity and reduces energy consumption. The system employs YOLO-based object detection optimized for fisheye distortion and demonstrates the practicality of responsible AI and edge computing for privacy-sensitive applications.

In Table~\ref{tab:capability_comparison} the main models are summarized and compared based on ten different features.
\begin{table*}[t]
\centering
\caption{Capability Comparison Among Privacy-Preserving Models}
\label{tab:capability_comparison}
\renewcommand{\arraystretch}{1.15}
\setlength{\tabcolsep}{4pt}
\begin{tabular}{p{3.6cm}p{1.25cm}p{1.25cm}p{1.25cm}p{1.25cm}c}
\hline
\textbf{Capability} & 
\textbf{\cite{popescu2022obfuscation}} & 
\textbf{\cite{jiang2023dartblur}} & 
\textbf{\cite{huang2024identity}} & 
\textbf{\cite{ghaziamin2024privacy}} & 
\textbf{Proposed model } \\ 
\hline
Operates in Image Space & $\checkmark$ & $\checkmark$ & $\checkmark$ & $\checkmark$ & $\times$ \\ 
Operates in Semantic/Textual Space & $\times$ & $\times$ & $\approx$ & $\times$ & $\checkmark$ \\ 
Face/Identity Protection & $\approx$ & $\checkmark$ & $\checkmark$ & $\checkmark$ & $\checkmark$ \\ 
Generalization to Non-Facial Data & $\checkmark$ & $\times$ & $\times$ & $\approx$ & $\checkmark$ \\ 
Adversarial Robustness & $\approx$ & $\checkmark$ & $\approx$ & $\approx$ & $\checkmark$ \\ 
Model Interpretability & $\times$ & $\times$ & $\approx$ & $\approx$ & $\checkmark$ \\ 
Privacy--Utility Balance & $\checkmark$ & $\checkmark$ & $\approx$ & $\approx$ & $\checkmark$ \\ 
Human Review Convenience & $\approx$ & $\checkmark$ & $\times$ & $\checkmark$ & $\checkmark$ \\ 
Scalability to Diverse Domains & $\approx$ & $\times$ & $\approx$ & $\approx$ & $\checkmark$ \\ 
Formal Privacy Guarantee (e.g., DP or abstraction) & $\times$ & $\times$ & $\times$ & $\times$ & $\checkmark$ \\ 
\hline
\end{tabular}
\end{table*}

\section{Proposed Solution}
\label{Sec:Method}

\begin{figure}
\centerline{\includegraphics[width=0.75\linewidth]{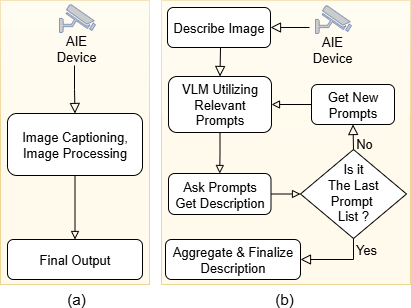}}
\caption{(a) General models follow a straightforward process which captures an image performing the image processing task and then produces the output.
(b) The proposed architecture begins with image capture and generates textual descriptions through an iterative cycle, using new prompts in each iteration.}
\label{fig:flowchart}
\end{figure}

This section presents the proposed hierarchical framework which integrates VLM, RL, and RAG to address image privacy challenges in ITS. 
Existing devices capture and process images through a linear pipeline. In contrast, we propose an iterative process which combines VLM and RL, supported by a feedback mechanism that generates new textual descriptions at each iteration. More specifically, 
visual data is first captured by AIE cameras and transformed into textual descriptions using a VLM, which generates responses based on selected prompts aimed at minimizing exposure of sensitive content. A feedback-driven RL agent is employed to iteratively refine the prompt selection process. However, since the VLM's prompt selection may not always generate optimal results, a RAG module is incorporated to monitor performance and enhance overall system effectiveness by retrieving and integrating relevant contextual information. The main components of the proposed framework are elaborated in the following four steps.


\begin{figure*}
\centerline{\includegraphics[width=\linewidth]{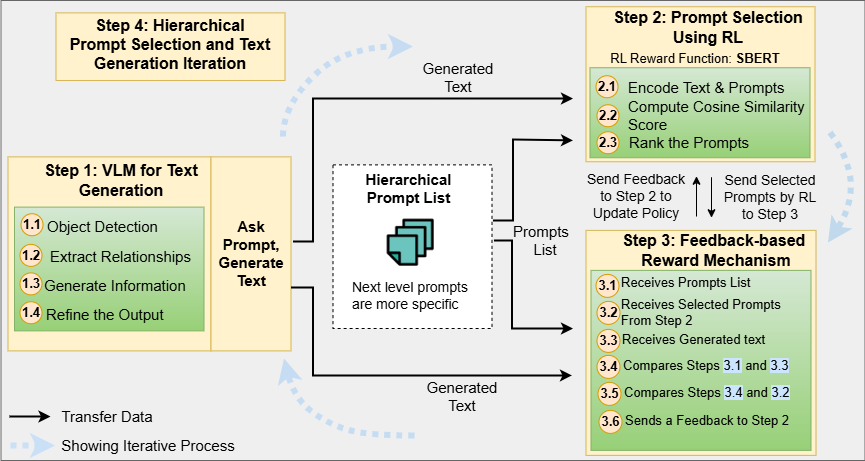}}
\caption{Feedback-based RL-VLM model running on AIE devices to capture images and transform into textual description through in an iterative process to generate maximum descriptions}
\label{RL_VLM}
\end{figure*}

\textbf{Step 1: VLM for Text Generation}

At the heart of the model is a VLM model that converts images into textual descriptions. This component is particularly effective due to its dual-modality architecture, which enables the model to learn joint representations of visual and linguistic information, as illustrated in Figure ~\ref{RL_VLM}. 

Unlike standard captioning models, the VLM is fine-tuned using privacy-sensitive supervision.
Images are paired with privacy-annotated captions (e.g., "driver visible", "face obscured") and penalized when text exposes private details.
A privacy classifier guides gradient updates so the VLM learns to avoid personally identifiable information (PII) descriptors during generation.
This adaptation aligns the latent visual–text representations toward privacy-aware semantics.

The VLM first performs object extraction with high accuracy, guided by the following loss function:

\begin{equation*}
     Loss = \sum_{i=0}^{s^2}\sum_{c\in classes} 1_{obj}^{ij}(p_i(c)- \hat{p_i}(c))^2
\end{equation*}

\noindent where $1_{obj}^i$ indicates whether an object exists in cell $i, p_i$ is the true class probability distribution for cell $i$, and $\hat{p}_i$ is the predicted distribution.

The VLM module generates descriptions by processing images based on prompts which allows it to produce rich summaries of visual content. These textual prompts are not arbitrary but rather follow a hierarchical structured list of predefined prompts that guide the model in extracting key visual elements, such as vehicles, pedestrians, or existing conditions within the image. This ensures that the descriptions are relevant and actionable for further processing. 
By encoding complex images into text, this step compresses the data while maintaining the semantic information necessary for an accurate representation in the cloud environment. 

Initially the model starts with token embedding 
$X = [x_1, x_2, ..., x_n], x_i \in \mathbb{R}^d$
where $X$ is the input matrix of token vectors, $n$ is the length of the sequence, and $d$ is the embedding dimension. Positional encoding are added to represent token positions in a sequence because through the following equation since Transformers do not have a built-in notion of sequence order:
$
    PE(pos,2i) = sin(\frac{pos}{10000^{\frac{2i}{d}}}), PE(pos, 2i+1) = cos(\frac{pos}{10000^{\frac{2i}{d}}})
$.
 
The positional embedding vector is added to the token embedding:
$
    Z_0 = X + P
$
where P represents the positional encoding. 
The weighted combinations of the token embeddings for each layer including input vectors are linearly transformed into $Query (Q)$, $Key (K)$, and $Value (V)$ matrices using learnable weight matrices $W^Q$, $W^K$, and $W^v$. 
The core of the attention mechanism is the scaled dot-product attention, which computes attention scores for each pair of tokens through:
$
    Attention(Q,K,V) = Softmax(\frac{QK^T}{\sqrt{d_k}})V
$
where $QK^T$ computed the dot-product similarity between queries and keys. The term $\frac{1}{\sqrt{d_k}}$ scales the dot product to prevent extremely large values, which could lead to small gradients. Also, the softmax normalizes these scores across all tokens, generating a probability distribution to determine the importance of each token. Then the self-attention mechanism is repeated multiple times in multi-head attention to learn different aspects of the input sequence: 
$
    MultiHeas(Q,K,V)=Concat(head_1,head_2,...,head_h)W^O
$
where each head is calculated using sets of weights, and $W^O$ is a weight matrix to linearly project the concatenated heads.
After the attention mechanism, each token's representation passes through a feed-forward network. This is done for each token independently, which consists of two linear transformations with a non-linear ReLU activation function:
$
    FFN(z)=ReLU(zW_1+b_1)W_2+b_2
$
where $W_1$, $W_2$ are weight matrices, and $b_1$, $b_2$ are biases.
To stabilize training, Layer Normalization and Residual Connections are used. 
In the first one each sub-layer (multi-head attention and FFN) has a skip connection that bypasses the layer and adds its input directly to the output. 
$
    Z^{'}_i=LayerNorm(Z_i+SubLayer(Z_i))
$
The Layer Normalization normalizes the inputs to stabilize and accelerate training, typically computed as:
$
    LayerNorm(z)=\frac{z-\mu}{\sigma+\epsilon}
$
where $\mu$ and $\sigma$ are the mean and standard deviation of the layer activations. 

The final output is generated using the softmax function to convert scores into a probability distribution over the vocabulary. This is computed for each token to predict the next token in the sequence:
$
    P(word_i) = Softmax(z_LW^V)
$
where $W^v$ is the learned weight matrix that projects the final token embeddings into vocabulary size. 
To train the model, backpropagation is used to update the weights to minimize the loss. The key calculations here include cross-entropy Loss function and a gradient descent optimization through the following equations:
$
    L=-\sum_{N}^{i=1}y_i log(\hat{y_i})
$
and 
$
    \theta_{t+1}=\theta_t-\alpha \frac{m_t}{\sqrt{v_t}+\epsilon}
$
where $m_{t+1}$ and $V_t$ are estimates of the first and second moments of the gradients, and $\alpha$ is the learning rate.

\begin{figure}
\centerline{\includegraphics[width=0.55\linewidth]{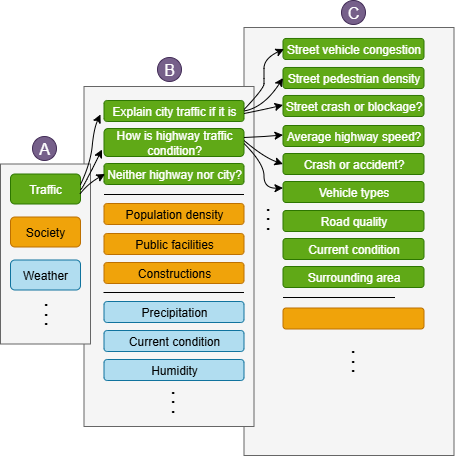}}
\caption{Hierarchical prompt selection where each layer prompts are selected at each iteration. As stated in Figure \ref{RL_VLM}, Step 3, the generated text is utilized to select the most relevant prompts from the prompt list during each iteration, where the list is updated in every cycle. }
\label{fig:hierarchical}
\end{figure}

\textbf{Step 2: Prompt Selection Using RL}

In this step, initial textual descriptions generated by the VLM are refined using the proposed RL model. 
While RL has been used in caption optimization, our contribution lies in integrating dual reward channels and hierarchical prompt refinement that iteratively enhance text relevance and privacy protection, an unexplored combination in current VLM literature.
The RL model optimizes the text by mapping it to the most relevant and specific prompts from a refined list. This process enhances both relevance and specificity which results in concise and context-aware descriptions that align with the requirements of the target digital environment. 
As shown in Figure~\ref{fig:hierarchical}, the model initially selects the most relevant prompts from list A, then performs the same process with list B, and finally with list C. This iterative selection continues as the generated text guides the subsequent levels of the hierarchy.

The RL model enhances hierarchical text generation by dynamically refining text based on training feedback, improving precision. It adjusts general descriptions by selecting prompts that highlight specific image details such as vehicles in traffic or industrial objects. Proximal Policy Optimization (PPO), a widely adopted RL method across various domains \cite{xue2024bidirectional}, is employed in this study to achieve the desired performance. The policy objective function for PPO is computed through the probability ratio:
\begin{equation*}
     r_{\theta}=\frac{\pi_{\theta}(a_{t}|s_{t})}{\pi_{\theta_{old}}(a_{t}|s_{t})}
\end{equation*}
where $r_\theta$ is the \textit{reward value}, $\pi_\theta$ is new \textit{policy}, $a_t$ is the \textit{action}, and $s_t$ is \textit{observation space}. 
The objective function is based on the \textbf{advantage function,} $A_{t}$, which estimates how much a new action is better than its older counterpart through the following expected reward:
\begin{equation*}
    L_t (\theta) = \mathbb{E} [r_t (\theta) A_t].
\end{equation*}

Considering PPO policy, the extended equation would be
\begin{equation*}
    L^{PPO} (\theta) = \mathbb{E}[min(r_\theta A_t , clip(r_\theta, 1 - \epsilon, 1 + \epsilon)A_t)]
\end{equation*}
which is comprised of $r_\theta = \frac{\pi_\theta(a_t|s_t)}{\pi_{\theta_{old}}(a_t|s_t)}$ as probability ratio, and $A_t$ as the advantage function $A_t = Q(s_t, a_t) - V(s_t)$. Considering all components, we get: 
\begin{equation*}
\begin{split}
    L^{PPO}(\theta)= \mathbb{E}[min(\frac{\pi_\theta(a_t|s_t)}{\pi_{\theta_{old}}(a_t|s_t)}).A_t, \\
    clip(\frac{\pi_\theta(a_t|s_t)}{\pi_{\theta_{old}}(a_t|s_t)}, 1-\epsilon,1+\epsilon).A_t]
\end{split}
\end{equation*}
If the new policy, controlled by $\epsilon$, improves the performance within a trust region then it updates normally. 
If the improvement is too large, the function prevents it from overshooting to avoid instability. This ensures stable learning and limits how much the new policy deviates from the old one per an update.

As illustrated in Figure \ref{RL_VLM}, after generating the refined text, a reward function is integrated into the core of the RL model. A pre-trained SBERT model \cite{reimers2019sentence} is employed to score and rank the refined prompts based on their semantic similarity to the input text. SBERT, a state-of-the-art model for generating sentence embeddings, enables efficient calculating of semantic similarity between textual elements.
In this process, SBERT encodes both the generated text and the refined list of prompts. Cosine similarity scores are then calculated between the text and each prompt. This similarity score represents how well the prompt aligns with the input text. Prompts are ranked based on these scores, allowing the model to filter out irrelevant or redundant information and pass only the most relevant prompts to the next stage.

\begin{algorithm}
\caption{The Proposed Algorithm with Feedback Mechanism}
\begin{algorithmic}[1]
\Require 
\State $iteration \geq 1$
\Ensure Policy is updated and refined prompts generated
\State $i \gets 1$
\State $iteration \gets 3$
\State $feedback \gets \textbf{False}$
\State $PU \gets \textbf{None}$
\While{$i \leq iteration$}:
    \State ObjectDetection()
    \If{$feedback$}
        \State UpdatePolicy($PU$)
        \State $feedback \gets \textbf{False}$
    \EndIf
    \State ComputeSimilarityScore()
    \State PromptSelection()
    \State ComputeReward()
    \State EvaluationAndSelection($Text$, $Prompt\ List$)
    \State UpdatePromptList()
    \State Compute $\mathcal{L}(\theta)$
    \State $s \gets \text{EvaluateSimilarity(): $armaxSim(q, D_i)$}$
    \State $PU \gets \text{PolicyUpdates(s)}$
    \If{$PU$}:
        \State $feedback \gets \textbf{True}$
    \EndIf
    \State $i \gets i + 1$
\EndWhile
\end{algorithmic}
\label{alg.steps}
\end{algorithm}

\textbf{Step 3: Feedback-Based Reward Mechanism}

Despite the successful extraction of the relative text from the images, the efficiency of the proposed model can be improved by adding an external function. The main idea is to propose a feedback-based RL framework to validate the process after extracting the refined text. To achieve this goal, RAG \cite{bag2024rag} is employed as an external knowledge augmentation mechanism to evaluate the model performance based on the generated feedback. Integrating  RAG into the RL agent enables the proposed model to not only provide effective feedback but also it improves the prompt management process, thereby improving the accuracy of the text refinement.

RAG offers advantages over PPO by dynamically retrieving the most relevant prompts which ensures contextual accuracy without the need for retraining. In contrast, PPO relies solely on RL and may struggle with new or evolving data. Moreover, RAG is more scalable and can efficiently operate over large knowledge bases.

To improve the accuracy of the generated text, the RAG model analyzes both the generated output and target prompts to identify the most relevant matches. This process occurs in parallel with the operation of the RL model. Once the RL agent selects the most relevant prompts, the RAG output is used to evaluate and validate the results.  
This stage is designed to correct potential errors or deficiencies in the RL model’s output and therefore supports improved convergence and overall performance.

The feedback-based process creates a loop in which the model iteratively improves its description generation by learning from these evaluations which enables it to produce highly relevant and precise descriptions.
The feedback mechanism operates based on the following equation:
\begin{equation*}
    L(\theta)={E}[r_t(\theta)A_t + \lambda.F(s_t,a_t)]
\end{equation*}
where $F(s,a)$ represents an external feedback function and $\lambda$ is a weighting coefficient that determines the influence of the feedback on the learning process.

In this context, the term $r_t(\theta)$ denotes the importance sampling ratio, given by $r_t(\theta)=\frac{\pi_\theta(a_t|s_t)}{\pi_{\theta_{old}}(a_t|s_t)}$. This ratio adjusts for the difference between the new and old policies. The advantage function, $A_t$, is defined as $A_t = Q(s_t, a_t) - V(s_t)$, which measures how much better an action $a_t$ is compared to the expected value under state $s_t$. Combining these definitions, the full loss function becomes:
\begin{equation*}
    \begin{split}        
L(\theta)=\mathbb{E_t}[(\frac{\pi_\theta(a_t|s_t)}{\pi_{\theta_{old}(a_t|s_t)}}).(Q(s_t,a_t)-\\
V(s_t))+ \lambda.F(s_t,a_t)]
    \end{split}
\end{equation*}

The augmented RL objective combines the classic policy gradient approach with external feedback. Specifically, $F(s_t,a_t)$ can represent a semantic similarity score such as one computed by RAG between the generated text and the ground truth. This integration enables the agent to learn not only from environment-based rewards but also from external, domain-specific signals. The hyperparameter $\lambda$ governs the trade-off between traditional policy learning and feedback correction, allowing flexible adjustment based on the quality and availability of external knowledge. This formulation improves the model's adaptability by integrating corrective feedback alongside environment-driven learning.

The RAG model's evaluation process is based on Maximum Inner Product Search (MIPS), where the most relevant document $\hat{D}$ is retrieved by computing $\hat{D}=argmaxSim(q,D_i)$, with $q$ denoting the query embedding generated by the encoder and $D_i$ representing the embedding of the $i-th$ document.

\textbf{Step 4: Hierarchical Prompt Selection and Text Generation Iteration}

The model uses a hierarchical structure to iteratively improve text descriptions by feeding selective prompts back to the text generator. This loop is repeated three times, progressively enhancing the accuracy and specificity of the descriptions, with a focus on the most important visual details.

By refining the output step by step, the system keeps only the essential information, removing redundant or irrelevant content. This not only reduces the volume of data transmitted to data centers but also ensures that each message precisely captures the corresponding visual scene. 






\subsection*{Reinforcement Learning Framework Details}

To ensure clarity of the technical design, we elaborate the RL framework which coveres the state and action space formulations, reward function, exploration strategy, and training configuration. 
The RL agent operates in conjunction with the VLM and RAG module, guiding the text refinement process toward privacy-aware and semantically precise descriptions.

\subsubsection*{a) State Space Definition}
The \textbf{state space} $S$ represents the environment information perceived by the RL agent at each iteration $t$. 
Each state $s_t$ encodes the semantic and privacy characteristics of the current textual output generated by the VLM:
\[
s_t = [E_{VLM}(x_t),\, P_{score}(x_t)]
\]
where $E_{VLM}(x_t)$ denotes the embedding vector obtained using the SBERT model for the generated text $x_t$, and $P_{score}(x_t)$ is a privacy indicator derived from the dissimilarity between the original and reconstructed visual content (i.e., the inverse of visual similarity). 
This feature representation enables the agent to evaluate the text both semantically and in terms of privacy leakage potential.

\subsubsection*{b) Action Space Definition}
The \textbf{action space} $A$ is defined as a discrete set of prompt-selection indices:
\[
A = \{a_1, a_2, ..., a_N\}
\]
Each action $a_t \in A$ corresponds to selecting one textual prompt from the refined prompt list that guides the VLM to generate a new descriptive variant. 
Once the action is taken, the VLM produces a refined description $x_{t+1}$, and the environment transitions to a new state $s_{t+1}$ accordingly.

\subsubsection*{c) Reward Function Design}
A composite reward function $R_t$ is designed to jointly optimize semantic privacy protection. 
The primary reward combines semantic similarity (measured by SBERT) and visual privacy (measured by reconstruction similarity):
\[
R_t = \alpha \cdot Sim_{SBERT}(x_t, x_{gt}) - \beta \cdot Sim_{VAE}(I_{orig}, I_{rec})
\]
where $Sim_{SBERT}$ denotes the cosine similarity between the generated text $x_t$ and ground-truth or contextual text $x_{gt}$, 
and $Sim_{VAE}$ denotes the similarity between the original image $I_{orig}$ and its reconstructed version $I_{rec}$, estimated via a variational autoencoder (VAE). 
The coefficients $\alpha$ and $\beta$ control the trade-off between relevance and privacy.

To incorporate knowledge-based feedback, the reward is augmented with an external feedback term from the RAG module:
\[
R'_t = R_t + \lambda \cdot F(s_t, a_t)
\]
where $F(s_t, a_t)$ represents the contextual validation score computed by RAG which indicates how strongly the retrieved information supports the generated text. 
The weighting factor $\lambda$ determines the influence of this feedback on the policy update.

\subsubsection*{d) Exploration–Exploitation Strategy}
Proximal Policy Optimization (PPO) is employed to ensure stable policy learning while maintaining a balance between exploration and exploitation. 
The clipped surrogate objective function is given by:
\[
L_{PPO}(\theta) = 
\mathbb{E}_t \left[ 
\min \left( r_t(\theta) A_t, 
\text{clip}(r_t(\theta), 1-\epsilon, 1+\epsilon) A_t 
\right) 
\right]
\]
where $r_t(\theta) = \frac{\pi_\theta(a_t|s_t)}{\pi_{\theta_{old}}(a_t|s_t)}$ is the probability ratio between new and old policies, and $A_t$ is the advantage function defined as $A_t = Q(s_t, a_t) - V(s_t)$. 
A decaying entropy term is added to encourage sufficient exploration during early training:
\[
L_{total}(\theta) = L_{PPO}(\theta) + \eta \cdot H[\pi_\theta(a|s)]
\]
where $H[\pi_\theta(a|s)]$ denotes the policy entropy and $\eta$ is gradually reduced (from $0.01$ to $0.001$) every 50 epochs. 
The clipping parameter $\epsilon$ is set to $0.2$ to maintain stable updates and prevent overshooting.

\subsubsection*{e) Network Architecture and Hyperparameters}
The policy and value networks are implemented using two fully connected layers of sizes $512$ and $256$, followed by ReLU activations. 
The output layer corresponds to the size of the action space $|A|$. 
The optimizer is Adam with a learning rate of $3\times10^{-5}$, and training proceeds for $500$ iterations using a batch size of $32$. 
Convergence is determined when both the cumulative reward and the semantic similarity score stabilize. 
The training pipeline is summarized as follows:
\begin{itemize}
    \item Initialize VLM parameters and prompt list.
    \item Encode initial text and compute: \\ $s_t = [E_{VLM}(x_t), P_{score}(x_t)]$.
    \item Select action $a_t$ based on policy $\pi_\theta(a|s_t)$.
    \item Generate refined text $x_{t+1}$ and compute $R_t$, $R'_t$.
    \item Update policy using $L_{total}(\theta)$.
\end{itemize}
This configuration ensures stable convergence while progressively improving both privacy protection and semantic richness of generated descriptions.

Algorithm \ref{alg.steps} outlines steps 1-4, demonstrating how the model begins with a base policy and iteratively improves it over three iterations. In each iteration, the policy is first updated based on similarity scores, followed by a second update driven by the computed loss function $L(\theta)$, applied when the change leads to measurable performance gains.

\section{Results and Discussions}
\label{Sec:Result}


The experimental framework is implemented in Python 3.11, where input images are processed to generate descriptive text outputs. To evaluate the generalizability of the proposed model, two datasets are employed: 
CFP-FP \cite{cfp-paper} dataset which is a face verification benchmark containing celebrity images in both frontal and profile views, designed to evaluate pose-invariant face recognition performance. Additionally, AgeDB-30 which is used for face verification under large age gaps, typically involving 570 identities and ~12,240 images.
The model will be compared with the results obtained from a VLM-based, RL-based, and feedback-based RL model then it will be compared to the model introduced in \cite{wang2023privacy} and \cite{peng2024fldatn}. 

The evaluation process involves four main categories of evaluations: 
\textit{Privacy Preservation Metrics},
\textit{Privacy Quantification and Risk Analysis},
\textit{Text Quality and Semantic Richness Metrics},
\textit{Ablation Study}. 

\begin{figure}
\centerline{\includegraphics[width=0.8\linewidth]{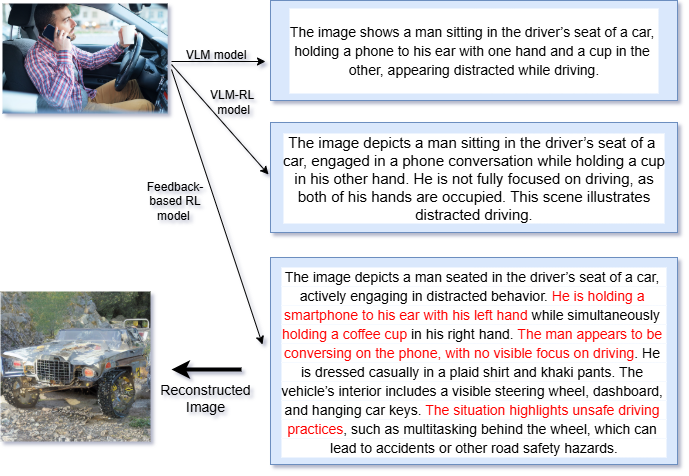}}
\caption{The figure illustrates sample textual descriptions generated by the proposed model and two additional models. The first description is produced by a VLM-based model, while the second is generated by a VLM-RL model. The third description is produced by the proposed model, which is capable of generating significantly longer and more detailed textual outputs. Additionally, the red-colored highlights indicate textual richness, showing that the proposed model effectively captures the main elements of the image. Using the generated text, an text to image model was used to evaluate how close is the reconstructed image to the input image in terms of information leakage. It is obvious that the re-generated image is a vehicle but without any similarity to the input image, meanwhile safety.}
\label{fig:result_flow}
\end{figure}

A sample textual description generated by the proposed model is shown in Figure~\ref{fig:result_flow} which is compared with the output from two other models. Then the textual description generated from the proposed model is used to generate a new image and it is shown to be significantly different from the input image.

\subsection{Privacy Preservation Metrics}
\begin{figure}
\centerline{\includegraphics[width=16pc]{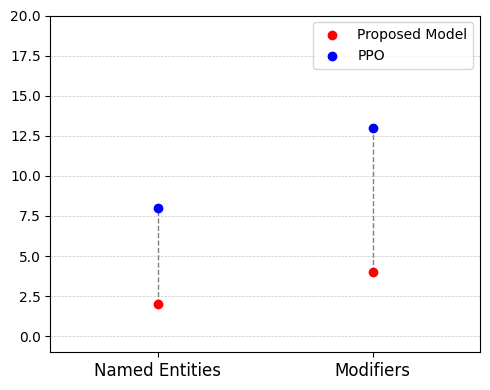}}
\caption{NER and Modifiers Metrics.}
\label{fig:detailed}
\end{figure}

To check how well privacy is protected, we use a test called \textit{the Adversarial Reconstruction Test}. In this test, pictures are reconstructed and then compared against the original pictures. The following metrics are used where Table \ref{tab:privacy_cfp} and \ref{tab:privacy_agedb} presents results for the first four metrics, metrics in 5 are shown in Figure~\ref{fig:detailed}: 
\begin{enumerate}
    \item SSIM evaluates how similar two images are.
    \item PSNR metric evaluates the quality of a compressed or reconstructed image by calculating the ratio between the maximum possible signal power and the power of the corrupting noise.
    \item MSE calculates how much each pixel in the reconstructed image differs from the corresponding pixel in the original image. 
    \item SRRA to measure the performance of the protected features.
    \item Named Entities and Modifiers evaluates whether the model correctly captures specific entities (e.g., people, places, objects) and uses rich descriptive modifiers. As illustrated in Figure ~\Ref{fig:detailed}, the baseline model often leaves such elements, while the proposed model consistently considers more. 
\end{enumerate}

\begin{table}[h!]
\centering
\caption{Privacy Metrics Evaluation on the CFP-FP Dataset }
\label{tab:privacy_cfp}
\begin{tabular}{@{}lcccc@{}}
\toprule
\textbf{Method} & \textbf{SSIM} $\downarrow$ & \textbf{PSNR} $\downarrow$ & \textbf{MSE} $\uparrow$ & \textbf{SRRA (\%)} $\downarrow$ \\ \midrule
AdvFace~\cite{wang2023privacy}  & 0.89 & 23.54 & 314.7 & 13.01 \\
FLDATN~\cite{peng2024fldatn}   & 0.91 & 24.32 & 349.1 & - \\
PPO-based RL  & 0.86 & 23.02 & 333.7 & 12.5 \\
Feedback-based RL (Ours) & \textbf{0.85} & \textbf{22.18} & \textbf{365.2} & \textbf{11.25} \\
\end{tabular}
\end{table}

\begin{table}[h!]
\centering
\caption{Privacy Metrics Evaluation on the AgeDB-30 Dataset }
\begin{tabular}{@{}lcccc@{}}
\toprule
\textbf{Method} & \textbf{SSIM} $\downarrow$ & \textbf{PSNR} $\downarrow$ & \textbf{MSE} $\uparrow$ & \textbf{SRRA (\%)} $\downarrow$ \\ \midrule
AdvFace \cite{wang2023privacy}  & 0.87 & 22.91 & 347.8 & 14.12 \\
FLDATN~\cite{peng2024fldatn}  & 0.89 & 24.59 & 369.1 & - \\
PPO-based  & 0.86 & 21.98 & 355.5 & 13.03 \\
Feedback-based RL (Ours) & \textbf{0.83} & \textbf{21.88} & \textbf{389.1} & \textbf{12.63} \\ \bottomrule
\end{tabular}
\label{tab:privacy_agedb}
\end{table}

\begin{table}[h!]
\centering
\caption{Differential Privacy Trade-Off Between Privacy Budget and Text Quality}
\begin{tabular}{lccc}
\hline
\textbf{Noise Level $\sigma$} & \textbf{$\varepsilon$ (↓)} & \textbf{BLEU Score (↓)} & \textbf{Privacy Gain (\% ↑)} \\
\hline
0.0 & $\infty$ & 0.831 & 0 \\
0.1 & 5.3 & 0.782 & 9.1 \\
0.2 & 3.1 & 0.779 & 15.7 \\
0.3 & 2.2 & 0.756 & 23.4 \\
\hline
\end{tabular}
\label{tab:dp_tradeoff}
\end{table}

\subsection{Privacy Quantification and Risk Analysis}

To establish a rigorous and quantifiable evaluation of privacy preservation, the proposed Hierarchical Feedback RL-VLM framework is complemented by formal privacy metrics and empirical risk analysis. Differential Privacy (DP) evaluation is performed to measure the privacy leakage and verify that textual transformation suppresses identity-linked cues rather than just removing visual content.

\subsubsection*{Differential Privacy Analysis}

In order to provide a formal privacy guarantee, a DP mechanism is applied to the sentence-embedding layer of the VLM. 
Gaussian noise $\mathcal{N}(0,\sigma^2)$ is injected into each textual embedding vector before storage or transmission:
\[
\tilde{E}_{VLM}(x_t) = E_{VLM}(x_t) + \mathcal{N}(0,\sigma^2)
\]
According to the Gaussian mechanism, the perturbed embedding satisfies $(\varepsilon,\delta)$-DP, where
\[
\varepsilon = \frac{\Delta_2 f}{\sigma}, \quad \delta < 10^{-5}
\]
and $\Delta_2 f$ is the $\ell_2$-sensitivity of the embedding function. 
By varying the noise scale $\sigma$, the privacy utility trade-off can be measured. 
A higher $\sigma$ ensures stronger privacy (smaller $\varepsilon$) at the cost of reduced linguistic fidelity.

Table~\ref{tab:dp_tradeoff} illustrates that adding small Gaussian noise significantly reduces the privacy budget $\varepsilon$ with minimal degradation in text quality which establishes a measurable privacy utility balance.

\subsection{Text Quality and Semantic Richness Metrics}
\begin{figure}
\centerline{\includegraphics[width=3.2in ]{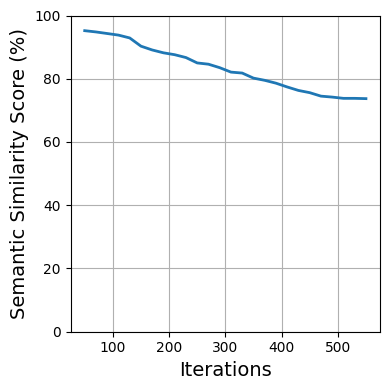}}
\caption{Semantic Similarity Score, which indicates how similar the text generated by the baseline method is similar to that of the proposed one.}
\label{fig:semantic}
\end{figure}

The information and linguistic quality of the generated text is assessed  by using the following literal and meaning metrics to evaluate the quality of privacy-preserved generated textual description: 
\begin{itemize}
    \item Word Count and Unique Word Count: reflect the length and literal variety of the output. As shown in Figure ~\Ref{fig:lexical}, the proposed model produces a longer and more detailed description than the baseline, and quantitatively indicates better expression.
    
    
    \item Semantic Similarity Score: compares two texts generated from the same image to assess the semantic similarity. The score of 0.7371 was obtained, confirming that the proposed model has high meaning alignment with baseline offering richer descriptive text. 

Figure \ref{fig:semantic} illustrates the semantic similarity between the text generated by the baseline model and the proposed model. Initially, the proposed model produces descriptions that are closely aligned with those of the baseline. However, as training progresses, the model receives updates and improvements. After 500 iterations, it reaches a stability, with the baseline model's output remaining approximately 73\% semantically similar to the text generated by the proposed model. This demonstrates that the proposed model is capable of enriching the generated text with additional semantic content.

\end{itemize}

\begin{figure}[t]
\centering
\begin{tikzpicture}
\begin{axis}[
    ybar,
    bar width=14pt,
    width=11cm,
    height=7cm,
    ymin=0,
    ymax=300,
    ylabel={Count},
    symbolic x coords={Word Count, Unique Word Count},
    xtick=data,
    nodes near coords,
    enlarge x limits=0.3,
    legend style={at={(0.5,1.05)}, anchor=south, legend columns=-1},
    ylabel style={font=\small},
    xlabel style={font=\small},
    tick label style={font=\small},
    label style={font=\small},
]
\addplot+[ybar, bar shift=-9pt, pattern=horizontal lines, pattern color=blue, fill=blue!30] 
coordinates {(Word Count,74) (Unique Word Count,46)};
\addplot+[ybar, bar shift=9pt, pattern=north east lines, pattern color=red, fill=red!30] 
coordinates {(Word Count,131) (Unique Word Count,89)};
\addplot+[ybar, bar shift=27pt, pattern=north east lines, pattern color=yellow, fill=yellow!30] 
coordinates {(Word Count,223) (Unique Word Count,149)};
\legend{VLM, RL-VLM, Proposed Method}
\end{axis}
\end{tikzpicture}
\label{fig:lexical}
\caption{Lexical Richness metrics count the number of words such the proposed method outperforms other models.}
\end{figure}


\subsection{Ablation and Comparative Study}

To understand the contribution of each component in the proposed hierarchical feedback-based RL-VLM model, an ablation study was conducted. 
Four model variants were evaluated by enabling (i) hierarchical RL and (ii) feedback-based retrieval using RAG. 
All models were tested on the same experimental setup described earlier.

\begin{table*}[h!]
\centering
\caption{Model Configurations Used in the Ablation Study}
\begin{tabular}{p{1.5in}ccc}
\hline
\textbf{Model Variant} & \textbf{VLM} & \textbf{Hierarchical RL} & \textbf{Feedback (RAG)} \\
\hline
{\small Baseline (VLM Only)} & \checkmark & $\times$ & $\times$ \\
{\small RL-VLM (Single - Stage)} & \checkmark & \checkmark & $\times$ \\
{\small RL-VLM + RAG (Feedback)} & \checkmark & \checkmark & \checkmark \\
{\small Full Model (Hierarchical Feedback RL-VLM)} & \checkmark\checkmark & \checkmark\checkmark & \checkmark\checkmark \\
\hline
\end{tabular}
\label{tab:model_variants}
\end{table*}

Figure~\ref{fig:ablation_pgflike} presents the quantitative evaluation across both privacy and text-quality metrics. 
Performance consistently improves as each component is integrated, demonstrating the significance of hierarchical learning and feedback refinement.

\subsubsection*{Qualitative Comparison}
Figure~\ref{fig:ablation_pgflike} illustrates the qualitative differences between the four variants for the same visual input. 
The VLM-only baseline produces generic one-line captions with limited privacy awareness, while the RL-VLM variant adds contextual structure but lacks external correction.
Integrating RAG feedback leads to semantically richer text and the full Hierarchical Feedback RL-VLM obtains the most descriptive and privacy-aware output.



\subsubsection*{Discussion}
The ablation results demonstrate that:
\begin{enumerate}
    \item \textbf{Hierarchical RL} substantially enhances semantic precision and fine-grained scene understanding by guiding prompt refinement across multiple iterations.
    \item \textbf{Feedback via RAG} introduces external semantic correction, improving linguistic diversity and alignment with privacy objectives.
    \item The \textbf{combined iterative framework} achieves the best overall performance, significantly outperforming traditional single-pass VLM captioning and single-stage RL baselines.
\end{enumerate}

Overall, the experiments validate that the proposed hierarchical feedback-based reinforcement learning framework extends beyond prior captioning and RL-VLM approaches, providing a systematic and adaptive mechanism for privacy-preserving visual-to-text transformation.

\begin{figure}[htbp]
\centering
\begin{tikzpicture}
\begin{axis}[
    ybar,
    bar width=7pt,
    width=\linewidth,
    height=6cm,
    enlargelimits=0.15,
    ylabel={Normalized Score (0--1)},
    symbolic x coords={
        SSIM ($\downarrow$),
        MSE ($\uparrow$),
        SRRA ($\downarrow$),
        Unique Words ($\uparrow$),
        Detail Density ($\uparrow$),
        Semantic Score ($\uparrow$)
    },
    xtick=data,
    x tick label style={rotate=25, anchor=east},
    ymin=0, ymax=1,
    legend style={at={(0.5,-0.38)},anchor=north,legend columns=3,draw=none},
    nodes near coords,
    every node near coord/.append style={font=\scriptsize, rotate=90, anchor=west},
    tick label style={font=\small},
    ylabel style={font=\small},
    grid=major,
    major grid style={dashed,gray!20}
]
\legend{VLM Only, RL-VLM, Full (HFR-VLM)}
\addplot+[ybar, bar shift=-9pt, pattern=horizontal lines, pattern color=black]
coordinates {
    (SSIM ($\downarrow$), 0.10)
    (MSE ($\uparrow$), 0.15)
    (SRRA ($\downarrow$), 0.35)
    (Unique Words ($\uparrow$), 0.22)
    (Detail Density ($\uparrow$), 0.40)
    (Semantic Score ($\uparrow$), 0.33)
};

\addplot+[ybar, bar shift=0pt, pattern=north east lines, pattern color=black]
coordinates {
    (SSIM ($\downarrow$), 0.3)
    (MSE ($\uparrow$), 0.23)
    (SRRA ($\downarrow$), 0.43)
    (Unique Words ($\uparrow$), 0.28)
    (Detail Density ($\uparrow$), 0.55)
    (Semantic Score ($\uparrow$), 0.62)
};

\addplot+[ybar, bar shift=9pt, pattern=dots, pattern color=black]
coordinates {
    (SSIM ($\downarrow$), 1.00)
    (MSE ($\uparrow$), 1.00)
    (SRRA ($\downarrow$), 1.00)
    (Unique Words ($\uparrow$), 1.00)
    (Detail Density ($\uparrow$), 1.00)
    (Semantic Score ($\uparrow$), 1.00)
    
};

\end{axis}

\end{tikzpicture}
\caption{Ablation Study Comparison of Model Variants. 
}
\label{fig:ablation_pgflike}
\end{figure}


\section*{Conclusion and Future Work}
This study introduces a privacy-preserving framework for ITS leveraging VLMs and RL. The proposed framework transforms visual data captured in ITS into concise, semantically rich text with the aim of preserving the individual privacy in the captured images. It demonstrates improved scalability, accuracy, and privacy protection, making it suitable for deployment in smart city infrastructures and industrial automation environments. Evaluations performed both on privacy-related and text quality metrics confirm the effectiveness of the proposed method. Although the generated text captures key elements of the image, it excludes sensitive visual details, thereby preserving user privacy. The method can produce three to four times more descriptive content compared to baseline models, without compromising privacy. For future work, we propose exploring dynamic prompt generation, where the prompt list is adaptively updated at each iteration based on context or feedback. This would enable more flexible, personalized, and privacy-aware text generation, further enhancing the model's generalizability and effectiveness.





\bibliographystyle{elsarticle-num}

\bibliography{References}





\end{document}